\documentclass[runningheads]{llncs}

\usepackage{eccv}

\usepackage{eccvabbrv}

\usepackage{graphicx}
\usepackage{booktabs}

\usepackage[accsupp]{axessibility}  

\usepackage{hyperref}

\usepackage{orcidlink}

\begin{document}

\title{Quantum-Gated LiteSSD: A Parameter-Efficient Lightweight Hybrid Quantum-Classical Framework for Forward-Looking Sonar Object Detection}
\titlerunning{Quantum-Gated LiteSSD}

\author{Niloy Kumar Mondal\inst{1} \and
Poulomi Sarker Puja\inst{2}}

\authorrunning{N.~K.~Mondal and P.~S.~Puja}

\institute{Department of Computer Science and Engineering,\\
Bangladesh University of Engineering and Technology, Dhaka 1000, Bangladesh\\
\email{2105044@ugrad.cse.buet.ac.bd} \and
Department of Urban and Regional Planning,\\
Khulna University of Engineering \& Technology, Khulna 9203, Bangladesh\\
\email{puja2117005@stud.kuet.ac.bd}}

\maketitle

\begin{abstract}
Forward-looking sonar object detection is essential for underwater
perception, yet deployment on embedded platforms requires highly compact
models. To address this challenge, we explore quantum computing and introduce
\textbf{Quantum-Gated LiteSSD}, a parameter-efficient hybrid
quantum--classical detector that reformulates QuCNet-style multi-circuit
quantum processing as an identity-centred channel-gating mechanism for spatial
feature modulation. Experiments on the Marine Debris Watertank dataset and UATD
forward-looking sonar benchmarks demonstrate an effective
parameter--accuracy trade-off. The proposed detector achieves 90.84\%
$\mathrm{mAP}_{50}$ on Watertank with approximately $62\times$ fewer
parameters than YOLO26s and $164.3\times$ fewer than SSD--VGG16. On UATD, the model achieves 70.37\% $\mathrm{mAP}_{50}$ with only
0.150M parameters, making it approximately $4.1\times$ smaller than
SSGA-YOLO while retaining meaningful multi-class detection capability.

\keywords{Forward-looking sonar \and Underwater object detection \and
Hybrid quantum--classical learning \and Parameter-efficient detection}
\end{abstract}

\section{Introduction}
\label{sec:intro}

Underwater object detection is essential for autonomous inspection,
environmental monitoring, and search-and-recovery operations. However,
degraded visibility, complex backgrounds, weak target boundaries, and
sensor noise make reliable recognition considerably more difficult than
in conventional terrestrial imagery~\cite{jian2024underwater,zhang2025enhancing}.
Moreover, the limited power, memory, and computational capacity of AUVs
and ROVs motivate lightweight detectors that balance detection accuracy
with real-time onboard deployment~\cite{wang2022ulo}.

QuCNet~\cite{komal2026qucnet}, a quantum machine learning (QML)
architecture, demonstrates that remote-sensing image classification can
retain competitive accuracy with roughly $10^{2}$--$10^{3}$ fewer trainable
parameters than representative classical networks. QML processes encoded
data using parameterized quantum circuits, commonly implemented as
variational quantum circuits, whose trainable gates exploit superposition
and entanglement to construct expressive nonlinear
representations~\cite{biamonte2017quantum,abbas2021power}. However, deep or
poorly designed circuits may exhibit vanishing gradients and barren
plateaus, making optimization difficult~\cite{larocca2025barren}. QuCNet
mitigates these trainability issues using multiple shallow parallel circuits
and highly compact
parameterization~\cite{komal2026qucnet}.

Despite this potential, parameter-efficient QML remains largely unexplored
for forward-looking sonar object detection, where the learned representation
must support both semantic discrimination and spatial localization. To address
this gap, we introduce \textbf{Quantum-Gated LiteSSD}, a parameter-efficient,
lightweight hybrid quantum--classical framework for forward-looking sonar
object detection. On the Watertank dataset, the proposed model achieves
90.84\% $\mathrm{mAP}_{50}$ with approximately $62\times$ fewer parameters
than YOLO26s and $164.3\times$ fewer parameters than SSD--VGG16. On UATD, the model achieves 70.37\% $\mathrm{mAP}_{50}$ with only
0.150M parameters, demonstrating non-trivial detection performance while
using approximately $4.1\times$ fewer parameters than SSGA-YOLO.

\section{Related Work}
\label{sec:related}
Recent forward-looking sonar object detectors have increasingly emphasized
computational efficiency and deployability on resource-constrained underwater
platforms. SSGA-YOLO~\cite{liu2026ssga} addresses this objective through
efficient convolutional operations and acoustic-aware attention, but remains a
fully classical architecture. Li and Ghosh~\cite{li2020quantum} applied quantum optimization to
bounding-box suppression through a Quadratic Unconstrained Binary
Optimization (QUBO) formulation; however, the quantum component operates
only as a non-learnable post-processing stage and does not participate in
feature learning or end-to-end detector training. Existing quantum learning studies for underwater
sensing have primarily focused on classification; for example, Bach
\etal~\cite{bach2026quantum} classify passive-sonar ship-propeller signals using
a hybrid quantum--classical model. QDCKN~\cite{kumaran2026novel} considers
marine object detection in optical underwater imagery, but employs
quantum-inspired operations rather than trainable parameterized quantum
circuits or quantum hardware. Similarly, QYOLO~\cite{mittal2026qyolo} improves
lightweight object detection through quantum-inspired shared channel mixing,
while remaining a fully classical model without executable quantum circuits. Q-Seg~\cite{venkatesh2024q} applies quantum annealing to unsupervised image segmentation, but formulates segmentation as a discrete optimization problem rather than learning trainable quantum features for object detection. QuCNet~\cite{komal2026qucnet} demonstrates
parameter-efficient trainable quantum circuits for remote-sensing image
classification, yet it is not designed to preserve spatial representations
required for object localization. Roh \etal~\cite{roh2024fast} designed a quantum convolution-based detector for
low-complexity autonomous-driving scenarios, whose direct transfer to
challenging underwater environments is limited by architectural choice. To the best of our knowledge, no prior work
has integrated trainable quantum circuits into a parameter-efficient
forward-looking sonar object detector.

\section{Proposed Quantum-Gated LiteSSD Architecture}
\label{sec:architecture}

We propose \textbf{Quantum-Gated LiteSSD}, a compact hybrid
quantum--classical detector for forward looking sonar images. The
architecture combines selected components adopted from
QuCNet~\cite{komal2026qucnet} with a newly designed quantum-gated SSD
detection pipeline. We adopt three main ideas from QuCNet~\cite{komal2026qucnet}.
First, a lightweight scaled feature extraction strategy is used to
obtain a compact classical representation. Second, the extracted
features are processed using sixteen parallel four-qubit trainable
quantum circuits. Third, Hybrid Cyclic Weight Sharing is used so that
the sixteen circuits reuse four groups of trainable parameters. Unlike QuCNet, our architecture uses the quantum outputs to control
the channels of a spatial feature map. This is the main architectural
difference and enables the quantum module to be used for object
localization.

Given a grayscale sonar image
$\mathbf{I}\in\mathbb{R}^{1\times300\times300}$, three SFE
blocks~\cite{komal2026qucnet} progressively transform the input dimension as
$1\times300\times300 \rightarrow 16\times150\times150
\rightarrow 32\times75\times75 \rightarrow 64\times38\times38$.
Let $\mathbf{F}\in\mathbb{R}^{B\times64\times38\times38}$ denote the
resulting feature map. Global average pooling is applied to each channel,
where $d_{b,c}=\frac{1}{38^{2}}\sum_{h=1}^{38}\sum_{w=1}^{38}
F_{b,c,h,w}$, producing the descriptor
$\mathbf{d}\in\mathbb{R}^{B\times64}$. The descriptor is then divided
into sixteen four-dimensional groups
$\mathbf{d}^{(j)}\in\mathbb{R}^{4}$, each of which is processed by one
four-qubit circuit. Before quantum encoding, the inputs are normalized
as $\boldsymbol{\phi}^{(j)}=\pi\tanh(\mathbf{d}^{(j)})$.

After the trainable quantum transformations, the $j$-th circuit outputs
a probability vector $\mathbf{p}^{(j)}\in\mathbb{R}^{16}$. A shared
linear projection computes
$\mathbf{u}^{(j)}=\mathbf{W}_{Q}\mathbf{p}^{(j)}+\mathbf{b}_{Q}$, where
$\mathbf{W}_{Q}\in\mathbb{R}^{4\times16}$. The projected outputs of all
sixteen circuits are concatenated to form
$\mathbf{u}=[\mathbf{u}^{(1)},\ldots,\mathbf{u}^{(16)}]
\in\mathbb{R}^{64}$. The proposed identity-centred channel gate is then
defined as $\mathbf{g}=1+\tanh(\mathbf{u})$, and the gated feature map is
obtained as
$\widetilde{\mathbf{F}}_{b,c,h,w}=g_{b,c}F_{b,c,h,w}$. Therefore, each
channel can be suppressed, preserved, or amplified while the original
$38\times38$ spatial layout remains unchanged for localization.

The quantum-gated feature map forms the first SSD feature level
$\mathbf{P}_{3}\in\mathbb{R}^{64\times38\times38}$. Four lightweight
depthwise-separable downsampling blocks and one adaptive-pooling block
generate the remaining levels:
$\mathbf{P}_{4}\in\mathbb{R}^{96\times19\times19}$,
$\mathbf{P}_{5}\in\mathbb{R}^{128\times10\times10}$,
$\mathbf{P}_{6}\in\mathbb{R}^{128\times5\times5}$,
$\mathbf{P}_{7}\in\mathbb{R}^{96\times3\times3}$, and
$\mathbf{P}_{8}\in\mathbb{R}^{64\times1\times1}$.
The Depthwise-separable classification and regression heads
predict class scores and box offsets, optimized end to end by
$\mathcal{L}_{\mathrm{cls}}+\mathcal{L}_{\mathrm{box}}$.
The quantum-gating module contains 132 trainable parameters: 64 quantum circuit angles and 68 classical parameters in the shared $16\!\rightarrow\!4$ projection.
\begin{figure}[tb]
    \centering
    \includegraphics[
        width=\textwidth,
        height=0.35\textheight,
        keepaspectratio
    ]{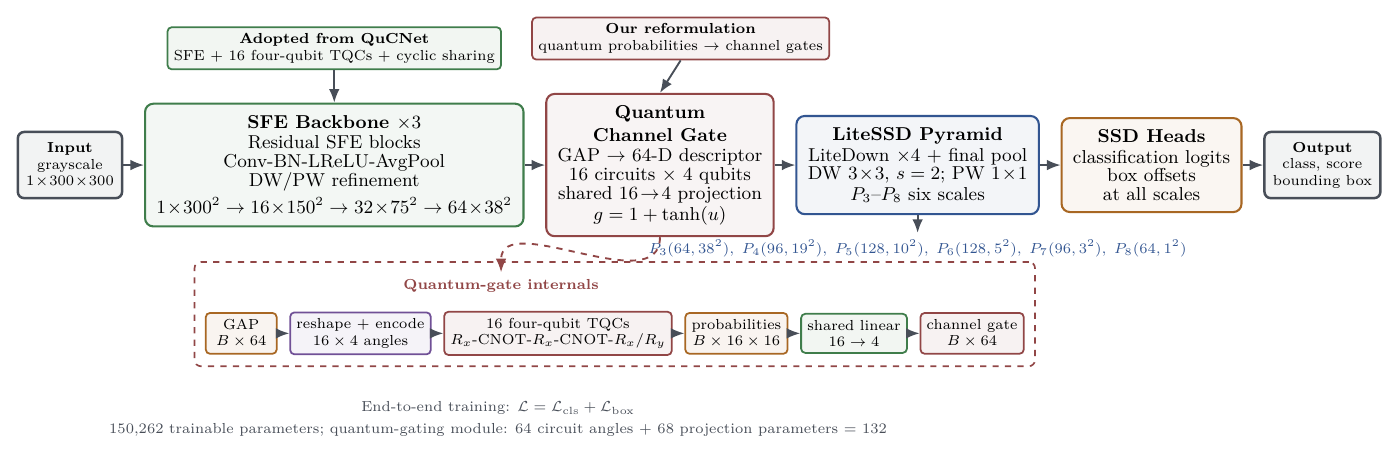}
    \caption{Module-level and end-to-end pipeline of the proposed
    Quantum-Gated LiteSSD detector.}
    \label{fig:qg_litessd}
\end{figure}

\section{Experimental Setup}
\label{sec:experimental_setup}

\noindent\textbf{Dataset, Preprocessing and Data Augmentation.}
We evaluate our quantum architecture on two publicly available forward-looking
sonar object-detection benchmarks: the Marine Debris Watertank
dataset~\cite{valdenegro2025marine} and the Underwater Acoustic
Target Detection (UATD) dataset~\cite{xie2022dataset}. The Watertank
object-detection subset contains 1,868 full-size grayscale sonar
images where the annotations contain 11 foreground categories: bottle,
can, chain, drink carton, hook, propeller, shampoo bottle, standing
bottle, tire, valve, and wall. We construct a deterministic
70/20/10 split, resulting in 1,308
training, 373 validation, and 187 held-out test images.
The UATD dataset contains 9,200 multibeam forward-looking sonar images across ten object categories~\cite{liu2026ssga}. We use the official split of 7,600 training images, 800 validation images (\texttt{UATD\_Test\_1}), and 800 held-out test images (\texttt{UATD\_Test\_2}).

All sonar images are converted to a single intensity channel,
normalized to the range $[0,1]$, and resized to
$300\times300$ pixels. During training, we apply horizontal
flipping with probability 0.5. Sonar-intensity augmentation is
performed using a random multiplicative gain sampled from
$[0.90,1.10]$ and an additive bias sampled from
$[-0.04,0.04]$. Gaussian noise with standard deviation 0.015 is
additionally applied with probability 0.25. No augmentation is
applied to the validation or test images.

\noindent\textbf{Evaluation Metrics.}
For the Watertank dataset, we report Pascal-VOC-style mean average
precision at an intersection-over-union threshold of 0.50,
denoted by $\mathrm{mAP}_{50}$. This metric is used to maintain consistency
with the SSD results reported by
Valdenegro-Toro et al.~\cite{valdenegro2025marine}. For UATD, we
use the COCO evaluation protocol and report
$\mathrm{mAP}_{50}$ and $\mathrm{mAP}_{50:95}$, where the latter
averages AP over IoU thresholds from 0.50 to 0.95 in increments of
0.05. 

\noindent\textbf{Implementation Details.}
QG-LiteSSD is implemented in PyTorch using the \texttt{torchvision} SSD interface with six-scale feature prediction and 16 four-qubit circuits simulated through an exact differentiable state-vector implementation. The model uses 64 trainable quantum angles and a 68-parameter shared projection, resulting in 153,300 parameters on Watertank and approximately 150,262 on UATD; all experiments are trained from scratch on an NVIDIA Tesla T4 using mixed precision.

\noindent\textbf{Training Configuration.}
On Watertank, we train for 120 epochs using AdamW, a batch size of 32, an initial learning rate of $2\times10^{-3}$, weight decay of $10^{-4}$, linear warm-up, and cosine decay, selecting the checkpoint with the highest validation $\mathrm{mAP}_{50}$. On UATD, we use Nesterov SGD for at most 120 epochs with a batch size of 128, learning rate of $10^{-2}$, momentum of 0.937, weight decay of $5\times10^{-4}$, cosine scheduling, and early stopping based on validation $\mathrm{mAP}_{50:95}$.

\noindent\textbf{YOLO Baseline.}
We fine-tune pretrained YOLO26s~\cite{jocher2026ultralytics} on the Watertank 70/20/10 split for up to 150 epochs using $640\times640$ inputs, a batch size of 16, and AdamW with cosine scheduling. The validation-selected checkpoint is evaluated on the held-out test set, while the UATD SSGA-YOLO results are taken directly from Liu et al.~\cite{liu2026ssga}.

\section{Results, Discussion, Limitations \& Future Work}

\noindent\textbf{Results on Marine
Debris Watertank Dataset. }
Table~\ref{tab:detection_comparison} demonstrates that Quantum-Gated
LiteSSD achieves a strong balance between detection accuracy and model
compactness. With only 0.1533M trainable parameters, our detector reaches
an $\mathrm{mAP}_{50}$ of 90.84\%. This is only 0.85 percentage points
below SSD--VGG16, despite using approximately $164.3\times$ fewer
parameters. Moreover, our model outperforms SSD--ResNet20 by 0.99
percentage points while being $29.5\times$ smaller, and substantially
outperforms the MobileNet-, DenseNet121-, SqueezeNet-, and
MiniXception-based SSD variants, although these models contain between
$21.9\times$ and $24.3\times$ more parameters.

YOLO26s obtains the highest $\mathrm{mAP}_{50}$ of 94.79\%, exceeding
our model by 3.95 percentage points. However, this improvement requires
approximately 9.50M parameters, making YOLO26s about $62\times$ larger
than Quantum-Gated LiteSSD. Thus, our architecture retains more than
95\% of the YOLO26s $\mathrm{mAP}_{50}$ while using only about 1.6\% of
its parameter count. These results indicate that the proposed
quantum-guided channel modulation can preserve competitive detection
performance under an extremely constrained parameter budget.

\begin{table}[tb]
\centering
\caption{Comparison with SSD-based object detectors and YOLO26s on the
Marine Debris Watertank dataset. The SSD mAP@50 results are reported
by Valdenegro-Toro et al.~\cite{valdenegro2025marine}. Since that work
does not report model sizes, the parameter counts of the SSD baselines
are approximated from publicly available implementations of the
corresponding backbone and detection architectures. The YOLO26s and
QG-LiteSSD results are obtained from our experiments on the held-out
test split.}
\label{tab:detection_comparison}
\resizebox{\columnwidth}{!}{%
\begin{tabular}{lccc}
\hline
\textbf{Architecture}
& \textbf{Parameters (M)}
& \textbf{mAP@50 (\%)}
& \textbf{Size Relative to Ours} \\
\hline

YOLO26s \cite{jocher2026ultralytics}
& $\approx 9.50$
& \textbf{94.79}
& $62.0\times$ \\

SSD--VGG16 \cite{simonyan2015very}
& $\approx 25.18$
& 91.69
& $164.3\times$ \\

SSD--ResNet20 \cite{he2016deep}
& $\approx 4.52$
& 89.85
& $29.5\times$ \\

SSD--MobileNet \cite{howard2017mobilenets}
& $\approx 3.73$
& 70.30
& $24.3\times$ \\

SSD--DenseNet121 \cite{huang2017densely}
& $\approx 3.50$
& 73.80
& $22.9\times$ \\

SSD--SqueezeNet \cite{iandola2016squeezenet}
& $\approx 3.49$
& 68.37
& $22.8\times$ \\

SSD--MiniXception \cite{arriaga2017real}
& $\approx 3.36$
& 71.62
& $21.9\times$ \\

\hline

\textbf{Quantum-Gated LiteSSD (Ours)}
& \textbf{0.1533}
& 90.84
& \textbf{$1.0\times$} \\

\hline
\end{tabular}%
}
\end{table}

\noindent\textbf{Results on UATD Dataset. }
As shown in Table~\ref{tab:uatd_comparison}, Quantum-Gated LiteSSD is
also the smallest model evaluated on UATD, containing only 0.150M
trainable parameters. It is approximately $4.1\times$ smaller than
SSGA-YOLO, $11.8\times$ smaller than YOLOv5n, and between
$16.3\times$ and $20.1\times$ smaller than the recent nano-scale YOLO
variants. Compared with EfficientDet, the proposed detector reduces
the parameter count by approximately $43.7\times$. This substantial
compression highlights the effectiveness of combining lightweight
depthwise-separable feature extraction with a highly parameter-efficient
quantum channel gate.

\begin{table}[tb]
    \centering
    \caption{Comparison with lightweight object detectors on the UATD dataset.
    Baseline results are the published values reported by Liu \etal~\cite{liu2026ssga},
    whereas Quantum-Gated LiteSSD is evaluated using our implementation.}
    \label{tab:uatd_comparison}
    \resizebox{\textwidth}{!}{%
    \begin{tabular}{lrrrr}
        \toprule
        \textbf{Method}
        & \textbf{Parameters}
        & \textbf{mAP$_{50}$ (\%)}
        & \textbf{mAP$_{50:95}$ (\%)}
        & \textbf{Size vs.\ Ours} \\
        \midrule
        EfficientDet
        & 6.56M
        & 88.80
        & 48.30
        & 43.7$\times$ \\

        SSD-MobileNet
        & 3.73M
        & 83.90
        & 47.60
        & 24.9$\times$ \\

        YOLOv5n
        & 1.77M
        & 92.80
        & 49.30
        & 11.8$\times$ \\

        YOLOv8n
        & 3.01M
        & 96.30
        & 56.00
        & 20.1$\times$ \\

        YOLOv9-t
        & 2.80M
        & 96.40
        & 56.30
        & 18.7$\times$ \\

        YOLOv10n
        & 2.70M
        & \textbf{96.60}
        & 56.50
        & 18.0$\times$ \\

        YOLOv11n
        & 2.58M
        & 96.50
        & 56.80
        & 17.2$\times$ \\

        YOLOv12n
        & 2.56M
        & 96.50
        & 57.30
        & 17.1$\times$ \\

        YOLOv13n
        & 2.45M
        & 96.40
        & \textbf{57.60}
        & 16.3$\times$ \\

        SSGA-YOLO
        & 0.61M
        & 94.80
        & 52.50
        & 4.1$\times$ \\

        \midrule
        \textbf{Quantum-Gated LiteSSD (Ours)}
        & \textbf{0.150M}
        & 70.37
        & 26.83
        & \textbf{1.0$\times$} \\
        \bottomrule
    \end{tabular}%
    }
\end{table}

The extreme reduction in model size is accompanied by an accuracy
trade-off on the more challenging UATD benchmark. Quantum-Gated
LiteSSD achieves an $\mathrm{mAP}_{50}$ of 70.37\% and an
$\mathrm{mAP}_{50:95}$ of 26.83\%, whereas SSGA-YOLO obtains 94.80\%
and 52.50\%, respectively. The larger performance gap under
$\mathrm{mAP}_{50:95}$ suggests that the compact detector experiences
greater difficulty producing tightly localized bounding boxes across
stricter IoU thresholds. Nevertheless, the model retains meaningful
multi-class detection capability using only one quarter of the
parameters of SSGA-YOLO and a small fraction of those required by the
remaining baselines.

\noindent\textbf{Overall Parameter--Accuracy Trade-off.}
The two benchmarks reveal complementary characteristics of the proposed
architecture. On Watertank, Quantum-Gated LiteSSD provides a particularly
favorable parameter--accuracy trade-off, matching or surpassing several
much larger SSD detectors and approaching the performance of YOLO26s.
On UATD, the accuracy gap is more pronounced, but the model establishes
an extreme compactness operating point that is not represented by the
existing baselines. Therefore, the principal advantage of
Quantum-Gated LiteSSD is not maximum detection accuracy in isolation,
but its ability to provide functional sonar object detection with only
approximately 0.15M parameters. This compact footprint makes the model
a promising candidate for further investigation on memory- and
compute-constrained underwater platforms.

\noindent\textbf{Limitations \& Future Work.} Current quantum devices remain constrained by noise and limited connectivity~\cite{preskill2018quantum}; future work will evaluate real-device robustness, scalability, and quantum-gate interpretability.
\bibliographystyle{splncs04}
\bibliography{main}
\end{document}